\documentclass[11pt]{article}
\usepackage{epsfig}

\begin{document}

\title{Zipf's law and the creation of musical context}

\author{Dami\'{a}n H. Zanette \\ {\it Consejo Nacional de Investigaciones
Cient\'{\i}ficas y T\'{e}cnicas} \\ {\it Instituto Balseiro, 8400
Bariloche, R\'{\i}o Negro, Argentina}}

\date{\today}

\maketitle

\begin{abstract}

This article discusses the extension of the notion of context
from linguistics to the domain of music. In language, the
statistical regularity known as Zipf's law --which concerns the
frequency of usage of different words-- has been quantitatively
related to the process of text generation. This connection is
established by Simon's model, on the basis of a few assumptions
regarding the accompanying creation of context. Here, it is shown
that the statistics of note usage in musical compositions are
compatible with the predictions of Simon's model. This result,
which gives objective support to the conceptual likeness of
context in language and music, is obtained through automatic
analysis of the digital versions of several compositions. As a
by-product, a quantitative measure of context definiteness is
introduced and used to compare tonal and atonal works.
\end{abstract}

\baselineskip .25in

\section{Introduction}

The appealing affinity between the cognitive processes associated
with music and language has always motivated considerable
interest in comparative research (Patel, 2003). Both music and
language are highly structured human universals related to
communication, whose acquisition, generation, and perception are
believed to share at least some basic neural mechanisms (Maess et
al., 2001).  The analysis of these concurrent aspects has
naturally lead to the attempt of extending concepts and methods
of linguistics to the domain of musical expression. Grammar,
syntax, and semantics have been discussed in the framework of
music from a variety of linguistically-inspired viewpoints
(Bernstein, 1973; Lerdahl and Jackendorf, 1983; Agawu, 1991;
Patel, 2003).  This approach, however, does not always take into
account the crucial difference of nature between the information
conveyed by music and language.  Consequently, such discussions
often remain at the level of a metaphoric parallelism. A
scientifically valuable comparative investigation of music and
language should begin by an accurate definition of common
concepts in both domains.

In this article, I explore the possibility of extending to the
domain of music a quantitative feature of language, related to
the frequency of word usage --namely, Zipf's law. The significance
of Zipf's law for language has resulted to be a controversial
matter in the past  (Simon, 1955; Mandelbrot, 1959). However, the
most successful explanation of Zipf's law --given by Simon's
model-- is based on linguistically sensible assumptions,
associated with the mechanisms of text generation and the concept
of context creation (Simon, 1955; Montemurro and Zanette, 2002;
Zanette and Montemurro, 2004). This supports the assertion that
Zipf's law is relevant to language. Moreover, since it involves a
quantitative property, an extension to the domain of music can,
in principle, be precisely defined.

Zipf's law has already been studied in music from a
phenomenological perspective, without reference to any possible
connection between linguistics and music theory (Boroda and
Polikarpov, 1988; Manaris et al., 2003). The main aim of this
article is to discuss Zipf's law as a by-product of the creation
of musical context, attesting the validity of extending the
assumptions of Simon's model to music. I begin by reviewing the
formulation of Zipf's law and Simon's model for language, with
emphasis in their connection with the concept of context. Then, I
discuss the extension of this concept to music. Finally, I show
with quantitative examples that Simon's model can be successfully
applied to musical compositions, which provides evidence of
analogous underlying mechanisms in the creation of context in
language and music. Context thus arises as a well-defined common
concept in the two domains.

\section{Zipf's law and Simon's model in language}

\begin{figure}[h]
\centerline{\psfig{file=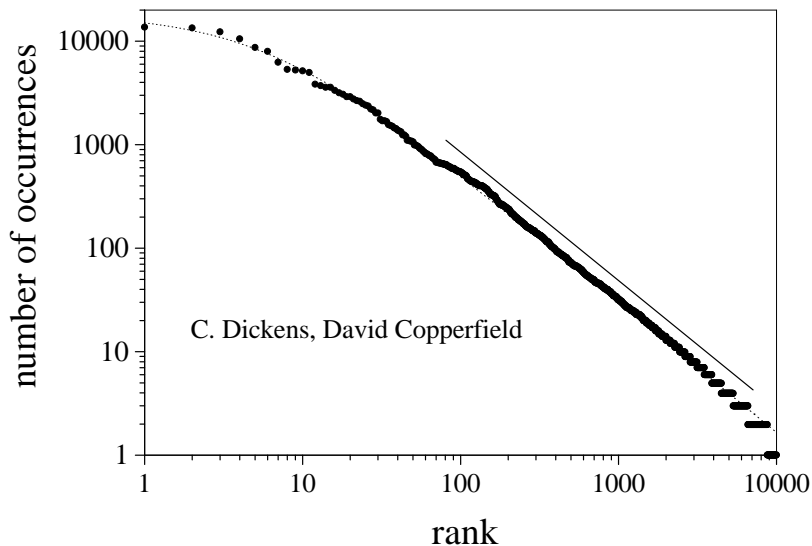,width=12 cm}} \caption{Zipf's
plot (number of occurrences $n$ versus rank $r$) for Dickens's
David Copperfield. The number of different words is $V=13,884$,
and the total number of words is $T=362,892$. In this
double-logarithmic plot the straight line manifests the power-law
dependence of $n(r)$ for large $r$. The dotted curve is a
least-square fitting with the prediction of Simon's model,
equation (\ref{eq}). } \label{fig1}
\end{figure}

In the early 1930s, G. K. Zipf pointed out a statistical feature
of large language corpora --both written texts and speech
streams-- which, remarkably, is observed in many languages, and
for different authors and styles (Zipf, 1935). He noticed that the
number of words $w(n)$ which occur exactly $n$ times in a
language corpus varies with $n$ as $w(n)\sim 1/n^\gamma$, where
the exponent $\gamma$ is close to $2$. This rule establishes that
the number of words with exactly $n$ occurrences decreases
approximately as the inverse square of $n$. Zipf's law can also be
formulated as follows. Suppose that the words in the corpus are
ranked according to their number of occurrences, with rank $r=1$
corresponding to the most frequent word, rank $r=2$ to the second
most frequent word, and so on. Then, for large ranks, the number
of occurrences $n(r)$ of the word of rank $r$ is given by
$n(r)\sim 1/r^z $, with $z$ close to $1$. The number of
occurrences of a word, therefore, is inversely proportional to
its rank. For instance, the $100$-th most frequent word is
expected to occur roughly $10$ times more frequently than the
$1000$-th most frequent word. Figure \ref{fig1} illustrates
Zipf's law for Charles Dickens's David Copperfield. All its
different words have been ranked, the number of occurrences $n$
of each word has been determined, and $n$ has been plotted
against the rank $r$. In this double-logarithmic plot, straight
lines correspond to the power-law dependence between $n$ and $r$
reported by Zipf.

Zipf himself advanced a qualitative explanation for the relation
between word frequency and rank, based on the balanced compromise
between the efforts invested by the sender and the receiver in a
communication process (Zipf, 1949). A quantitative derivation of
Zipf's law was later provided by H. A. Simon, in the form of a
model for text generation (Simon, 1955). The basic assumption
underlying Simon's model is that, as words are successively added
to the text, a context is created. As the context emerges, it
favours the later appearance of certain words --in particular,
those that have already appeared-- and inhibits the use of others.
In its simplest form, Simon's model postulates that, during the
process of text generation, those words that have not yet been
used are added at a constant rate, while a word that has already
appeared is used again with a frequency proportional to the
number of its previous occurrences.  These simple rules are
enough to prove that, in a sufficiently long text, the number
$w(n)$ of words with exactly $n$ occurrences is, as noticed by
Zipf, $w(n)\sim 1/n^\gamma$. The exponent $\gamma$ is determined
by the rate at which new words are added, and takes the observed
value $\gamma \approx 2$ when that rate is close to zero.

Simon's model can be refined by assuming that, as observed in
real texts, the rate of appearance of new words decreases as the
text becomes longer (Montemurro and Zanette, 2002; Zanette and
Montemurro, 2004).  Specifically, if the number $V$ of different
words varies with the length $T$ of the text as $V\sim T^\nu$,
with $0<\nu <1$, it turns out that $w(n)\sim 1/n^{1+\nu}$.
Assuming moreover that there exists an upper limit $n_0$ for the
number of occurrences of any single word, it is possible to show
that the number of occurrences as a function of the rank is
\begin{equation}
\label{eq} n(r) = \frac{1}{(a+br)^z}
\end{equation}
with $z=1/\nu$. The constants $a$ and $b$ are given in terms of
$n_0$ and $V$ as $a=1/n_0^\nu$ and $b=(1-1/n_0^\nu)/V$. The upper
limit $n_0$ is turn connected to $V$ and $T$ through the relation
$T/V=\nu (n_0^{1-\nu}-1)/(1-\nu)(1-1/n_0^\nu)$. For sufficiently
large ranks, the form of $n(r)$ given in equation (\ref{eq})
reproduces the expected ``Zipfian''  behaviour $n(r)\sim 1/r^z$.
The dotted curve in figure \ref{fig1} is a least-square fitting
of the data of $n$ vs $r$ for David Copperfield with equation
(\ref{eq}). The remarkable agreement between the data and the
fitting supports the hypotheses of Simon's model.

Thus, Simon's model interprets Zipf's law as a statistical
property of word usage during the creation of context, as a text
is progressively generated. As discussed in Section 3, the
creation of context in language is intimately related to the
semantic contents of words, i.e. to their meaning. Semantics is
in fact essential to the function of language as a communication
system. To assess the significance of Zipf's law and of the
assumptions of Simon's model in the framework of music one must
first examine to which extent the concepts of context and
semantic contents can be extended to musical expression.

\section{Semantics and context in music}

In contrast to language, music lacks functional
semantics.\footnote{Here, I use the word {\it semantics} in the
strict sense, namely, as the connection between symbols and the
entities that they represent in the extra-symbolic world.}
Generally, the musical message does not convey information about
the extra-musical world and, therefore, a conventional
correspondence between musical elements and non-musical objects
or concepts (i.e., a dictionary) is irrelevant to its cognitive
function. Unless music is accompanied by a text and/or by
theatrical action, its semantic contents is usually limited to the
onomatopoeic-like episodes of ``musical pictures'' or to a rather
rough outline of mood, frequently determined just by rhythm and
tonality. Assigning extra-musical meaning to a musical message is
basically an idiosyncratic matter, yielding highly non-universal
results.

On the other hand, the notion of context is essential to both
language and music. In the two cases, context can be defined as
the global property of a structured message that sustains its
coherence or, in other words, its intelligibility (van Eemeren,
2001).  Thus, such notion lies at the basis of the cognitive
processes associated with written and spoken communication and
with musical expression and perception. A long chain of words
--even if they constitute a grammatically correct text-- or a
succession of musical events --even if they form, for instance, a
technically acceptable harmonic progression-- would result
incomprehensible if they do not succeed at defining a contextual
framework. It is in this framework, created by the message
itself, that its perceptual elements become integrated into a
meaningful coherent structure.

In language, context emerges from the mutually interacting
meanings of words. As new words are successively added to a text
or speech stream, context is built up by the repeated appearance
of certain words or word combinations, by the emphasis on some
classes of nouns and adjectives, by the choice of tense, etc.
These elements progressively establish the situational framework
defined by the message in all its details. Thus, linguistic
context is a collective expression of the semantic contents of
the message.

In music, context is determined by a hierarchy of intermingled
patterns occurring at different time scales. For the occasional
listener, the most evident contribution to musical context
originates at the level of the melodic material, whose
repetitions, variations, and modulations shape the thematic base
of a composition (Schoenberg, 1967). The tonal and rhythmic
structure of melody phrases constitutes the substance of musical
context at that level. At larger scales, the recurrence of long
sections and certain standard harmonic progressions determine the
musical form. Crossed references between different movements or
numbers of a given work establish patterns over even longer
times. Meanwhile, at the opposite end of time scales, a few notes
are enough to determine tempo, rhythmic background, and tonality,
through their duration and pitch relations.

An obvious difficulty in modelling the creation of musical
context along the lines discussed in Section 2 for language,
which are based on the statistics of word usage, resides in the
fact that the notion of {\it word} cannot be unambiguously
extended to music (Boroda and Polikarkov, 1988). In language,
words --or short combinations of words-- stand for the units of
semantic contents, with (almost) unequivocal correspondence with
objects and concepts. Moreover, in the symbolic representation of
language as a chain of characters, i.e. as a written text, words
are separated by blank spaces and punctuation marks, which
facilitates their identification --in particular, by automatic
means. Music, on the other hand, does not possess any
conventionally defined units of meaning. The notion of {\it word}
is however conceivable in music by comparison with the linguistic
role of words as ``units of context,'' namely, as the perceptual
elements whose collective function yields coherence and
comprehensibility to a message. In music, the role of ``units of
context'' is played by the building blocks of the patterns which,
at different time scales, make the musical message intelligible.
Yet, the identification of such units in a specific work may
constitute a controversial task.

In the quantitative investigation of context creation in music, I
have chosen as ``units of context'' the building blocks of the
smallest-scale patterns, namely, single notes. A note is here
characterised by its pitch (i.e. its position on the clef-endowed
staff) and type (i.e. its duration relative to the tempo mark),
and its volume, timbre, and actual frequency and duration are
disregarded. The contribution of notes to the creation of musical
context, determining tonality and the basis for rhythm, is
particularly transparent. In addition, the choice of single notes
has several operational advantages. In the first place, the
collection of notes available to all musical compositions --or,
at least, to all those compositions that can be written on a
staff using the standard note types-- is the same. This
collection of notes plays the role of the lexicon out of which
the message is generated. Secondly, single notes are well-defined
entities in any symbolic representation of music, either printed
on a staff or in standardised digital formats, such as the
Musical Instrument Digital Interface (MIDI).  This makes possible
their automatic identification, which, as described later,
constitutes a crucial step in the analysis. Moreover, in order to
extract any meaningful information from a statistical approach,
it is necessary to work with relatively large corpora. The
compositions used in the present investigation contain, typically,
several thousand single notes. This figure remains well below the
number of words in any literary corpus, which usually reaches a
few hundred thousands (cf. figure \ref{fig1}), but is already
suited for statistical manipulations.

The convenience of choosing single notes as the ``units of
context'' is best appraised by comparing with other possible
choices. Consider, for instance, a definition of ``unit of
context'' in terms of melodic phrases. First of all, the limits
of a melodic phrase cannot be unambiguously determined.
Furthermore, unless one takes into account the infinitely vast
universe of all possible melodies, melodic phrases do not
constitute a common lexicon for different compositions. Finally,
since melodic phrases are subject to modulation and variation as
a work progresses, their automatic identification would demand
resorting to the sophisticated computational procedures.

\section{Application of Simon's model to music}

The starting point in the study of the relevance of Simon's model
to the creation of musical context, is Zipf's analysis of note
usage. I have employed a computational code to sequentially read
the MIDI version of a musical composition,\footnote{The MIDI
files of the musical compositions studied in this article are
available at www.geocities.com/benedetto{\_} marcello/midi/} and
detect the ``events'' corresponding to single notes. Each of these
``events'' consists of a sequence of hexadecimal digits, with
explicit information on the relative duration and pitch of the
corresponding note (Lehrman and Tully, 1993). This information is
extracted, and notes are ranked according to their number of
occurrences. I denote by $T$ the total number of notes (i.e. the
"text length," cf. Section 2) and by $V$ the number of different
notes (i.e. the "lexicon size").

I have performed Zipf's analysis on a variety of western music
works, from different periods, styles, and with different musical
forms. In this article, I present results for four compositions
for keyboard, which insures a certain degree of idiomatic
homogeneity in spite of the diversity of style. They are the
Prelude N.~6 in d from the second book of Das Wohltemperierte
Klavier, by J.~S.~Bach; the first movement, Allegro, from the
Sonata in C (K.~545) by W.~A.~Mozart; the second movement,
Menuet, from the Suite Bergamasque by C.~Debussy; and the first
of Three Piano Pieces (Op.~11, N.~1)  by A.~Schoenberg. In all
cases, I have disregarded short grace notes, which have not been
written down by the composer and whose realisation relies on the
performer, and have not taken into account full-section
repetitions, which contribute to musical context at the largest
time scales only.

\begin{figure}[h]
\centerline{\psfig{file=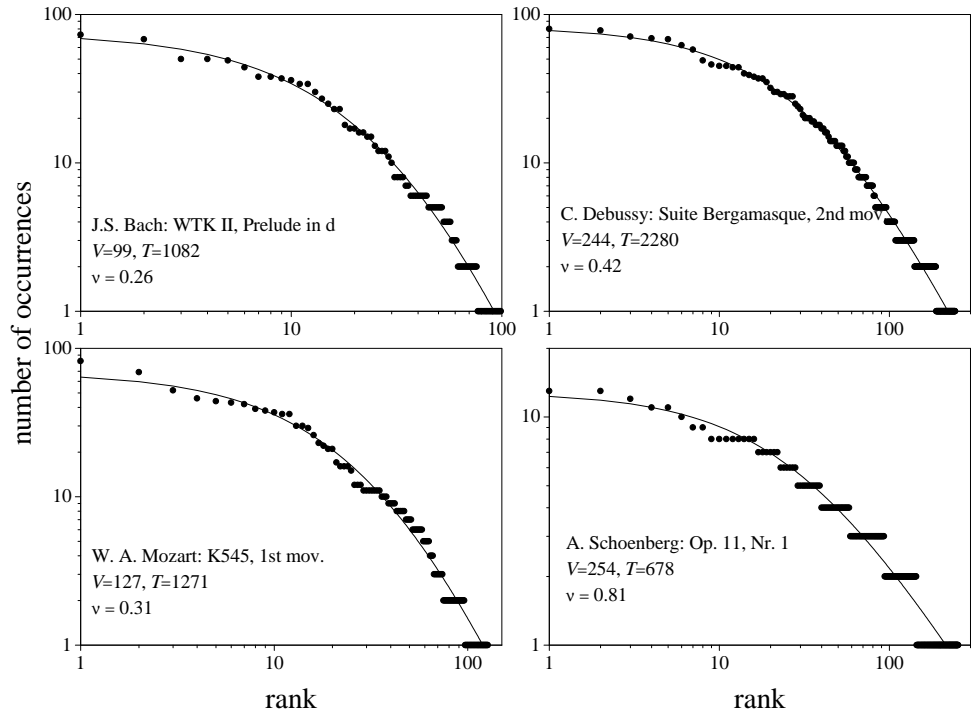,width=14 cm}} \caption{Zipf's
plots for single notes in four musical compositions for keyboard.
Their titles, as well as the corresponding value of $V$ and $T$,
are indicated in each panel. Curves stand for least-square
fittings with the prediction of Simon's model, equation
(\ref{eq}). The resulting exponent $\nu$, which provides a
quantitative measure of context definiteness, is given with each
plot.} \label{fig2}
\end{figure}

Figure \ref{fig2} shows, as full dots, the number of occurrences
$n$ versus the rank $r$ for single notes in the four works listed
above.  The respective values of $V$ and $T$ are indicated in
each panel.  The merest inspection of these Zipf's plots reveals
a striking similarity in the functional shape of $n(r)$ for the
four data sets.  I have obtained the same kind of shape for all
the compositions analysed following Zipf's prescription. This
similarity already suggests the existence of a common underlying
mechanism, determining the relative frequency at which different
notes are used, independent of work length, musical form,
tonality, style, and author.

Note that, in contrast to figure \ref{fig1}, the plots of figure
\ref{fig2} lack the long linear regime corresponding to the
power-law dependence of $n(r)$. This circumstance, which can be
ascribed to the relatively minute values of $V$ and $T$ for
musical compositions as compared with literary corpora, does not
preclude the application of Simon's model. In fact, according to
equation (1), the ``Zipfian'' regime is attained for sufficiently
large ranks only. The empirical data obtained from Zipf's
analysis of note usage must be rather compared with the full form
of $n(r)$, as given by equation (\ref{eq}).

Curves in figure \ref{fig2} stand for least-square fittings of the
data with equation (\ref{eq}). The constants $a$ and $b$ can be
calculated beforehand in terms of the respective values of $V$
and $T$, as discussed in section 2. Consequently, the only free
parameter to be determined by the fitting is the exponent $z$ or,
equivalently, the exponent $\nu=1/z$.  The resulting values of
$\nu$ are quoted in figure \ref{fig2}. The agreement between the
empirical data and the prediction of Simon's model is remarkably
good for the four data sets. A chi-square test of the quality of
fitness validates the hypothesis that these data are
statistically equivalent to equation (\ref{eq}) at a confidence
level close to $100$ {\%}.  This implies that the results of
Zipf's analysis are compatible with the hypothesis that
single-note usage follows the assumptions of Simon's model.
Specifically, they are in agreement with the assumption that the
occurrences of a given note promote its later appearance, with a
frequency that grows as the number of previous occurrences
increases. According to the above discussion, this process stands
for the basic mechanism of context formation.

While the four data sets shown in figure \ref{fig2} are
consistent with Simon's model and, in fact, display a common
functional dependence between $n$ and $r$, a quantitative
disparity between the four sets becomes apparent by comparing the
respective values of the exponent $\nu$, obtained from the
least-square fitting. Recall from Section 2 that this exponent
quantifies the functional relation between the lexicon size $V$
and the text length $T$, as $V\sim T^\nu$. Mathematically, $\nu$
can be identified with the ratio between a relative variation
$\Delta V/V$ in the lexicon size to the corresponding relative
variation $\Delta T/T$ in the text length. A small value for
$\nu$ corresponds a lexicon whose size increases slowly as
compared with the text growth, while a value close to one
corresponds to a lexicon growing at the same relative rate as the
text itself. Small exponents are therefore an indication of a
compact lexicon, determining a robust context that remains
relatively stable and well defined as the text progresses. On the
other hand, large exponents reveal an abundant lexicon, related
to a ductile, unsteady, more tenuously defined context. In terms
of context, therefore, the exponent $\nu$ can be interpreted as
quantitative measure of variability or, conversely, of
definiteness.

In the four musical works analysed here, the exponent $\nu$
happens to grow chronologically, following the composition dates.
Its variation from Bach to Debussy is however not significant. In
fact, the analysis of other keyboard works by Bach and Mozart
--for instance, other preludes from Das Wohltemperierte Klavier
and other sonatas-- yields values between $0.25$ and $0.45$. The
only significant difference corresponds therefore to Schoenberg's
Piano Piece. This work is well known as a landmark of consistent
atonality, where the construction of a tonal context has been
avoided on purpose (Perle, 1991). The absence of one of the
contextual elements determined at the level where single notes
act as ``units of context'' is clearly manifested by the large
value of $\nu$ resulting from the present analysis.

\section{Conclusion}

While the extension of the notion of semantic contents from
linguistics to music holds as a metaphoric allegory only, context
--whose role in language is closely related to semantics-- stands
for a significant feature common to linguistic and musical
messages. In both domains, context denotes a property emerging
from the interaction of the perceptual elements that compose the
message, that makes the message intelligible as a whole. The
nature of the information borne by music differs substantially
from that of language. However, the combination of those elements
in a hierarchically organised sequence, whose structure sustains
its comprehensibility, lies at the basis of the creation of
context in the two domains.

In this article, I have provided evidence supporting the assertion
that the definition of linguistic context can be shared with
music. Fortunately enough, context can be conceptually related to
a quantitative property of literary corpora, enunciated by Zipf's
law, whose validity in a musical corpus can be investigated by
objective means. It is Simon's model which establishes the
connection between message generation, context creation, and
Zipf's law. The evidence arises, therefore, from the confirmation
that musical corpora verify the predictions of Simon's model, an
approach that relies on purely mathematical operations. As a
by-product, this approach yields a quantitative measure of
context definiteness --the exponent $\nu$. A demonstration of this
measure has been drawn from the comparison of an atonal musical
work with tonal compositions: in the former, the absence of tonal
context results in a larger value of $\nu$.

Of course, the present mathematical approach is not
assumption-free. In particular, a crucial choice was made at the
moment of extending the notion of {\it word}to musical messages.
It would be interesting to consider alternative extensions, at
the level of melodic phrases, harmonic sequences, or rhythmic
patterns, and thus explore the concept of musical context at
different scales.

\section*{Acknowlegement}

I am grateful to M.~A.~Montemurro for his critical reading of the
manuscript.

\section*{References}

\noindent Agawu, V. K. (1991). Playing with Signs: A Semiotic
Interpretation of Classical Music. Princeton: Princeton
University Press. \vspace{10 pt}

\noindent Bernstein, L. (1973). The Unanswered Question.
Cambridge, MA: Harvard University Press. \vspace{10 pt}

\noindent Boroda, M. G. and Polikarpov, A. A. (1988). The
Zipf-Mandelbrot law and units of different text levels.
Musikometrika, 1, 127-158. \vspace{10 pt}

\noindent Lehrman, P. and Tully, T. (1993). MIDI for the
Professional. London: Music Sales.\vspace{10 pt}

\noindent Lerdhal, F. and Jackendorf, R. (1983). A Generative
Theory of Tonal Music. Cambridge, MA: MIT Press.\vspace{10 pt}

\noindent Maess, B.,  Koelsch, S., Gunter T., and Friederici, A.
D. (2001). Musical syntax is processed in Broca's area: an MEG
study. Nature Neurosciences, 4, 540-545.\vspace{10 pt}

\noindent Manaris, B., Vaughan, D., Wagner, C., Romero, J., and
Davis, R.B. (2003). Evolutionary music and the Zipf-Mandelbrot
law: Progress towards developing fitness functions for pleasant
music, in Lecture Notes in Computer Science: Applications of
Evolutionary Computing, pp. 522-534. Berlin:  Springer
Verlag.\vspace{10 pt}

\noindent Mandelbrot, B. B. (1959). A note on a class of skew
distribution function. analysis and critique of a paper by
H.~A.~Simon, Information and Control, 2,90-99.\vspace{10 pt}

\noindent Montemurro, M. A. and Zanette, D. H. (2002).  New
perspectives on Zipf's law in linguistics: From single texts to
large corpora. Glottometrics, 4, 86-98.\vspace{10 pt}

\noindent Patel, A. D. (2003). Language, music, syntax and the
brain. Nature Neurosciences, 6, 674-681.\vspace{10 pt}

\noindent Perle, G. (1991). Serial Composition and Atonality: An
Introduction to the Music of Schoenberg, Berg, and Webern.
Berkeley: University of California Press.\vspace{10 pt}

\noindent Schoenberg, A. (1967). Fundamentals of Musical
Composition. London: Faber and Faber.\vspace{10 pt}

\noindent Simon, H.A. (1955).  On a class of skew distribution
functions. Biometrika, 42, 425-440. Reprinted in Simon, H.A.
(1957). Models of Man. Social and Rational. New York: John Wiley
\& Sons.\vspace{10 pt}

\noindent van Eemeren, F. H. (2001). Crucial Concepts in
Argumentation Theory. Chicago: University of Chicago
Press.\vspace{10 pt}

\noindent Zanette, D. H. and Montemurro, M. A. (2004).  Dynamics
of text generation with realistic Zipf's distribution. Journal of
Quantitative Linguistics, in press.\vspace{10 pt}

\noindent Zipf, G. K. (1935). The Psycho-Biology of Language.
Boston: Houghton Mifflin.\vspace{10 pt}

\noindent Zipf, G. K. (1949). Human Behaviour and the Principle
of Least Effort. Cambridge, MA: Addison-Wesley.\vspace{10 pt}

\end{document}